\newif\ifarxiv
\arxivtrue 

\documentclass[letterpaper, 10 pt, conference]{ieeeconf}  

\IEEEoverridecommandlockouts                              
\usepackage[utf8]{inputenc}
\usepackage[T1]{fontenc}
\usepackage{amsmath}
\usepackage{mathrsfs}
\usepackage{amsfonts}
\usepackage{algorithm}
\usepackage{algorithmicx}
\usepackage{algpseudocode}

\usepackage{color}
\usepackage{bm}
\usepackage{diagbox}

\usepackage{graphicx}

\title{\LARGE \bf
Quasistatic contact-rich manipulation via \\ linear complementarity quadratic programming
}

\author{Sotaro Katayama$^{\dagger, 1}$, Tatsunori Taniai$^{2}$, and Kazutoshi Tanaka$^{2}$
\thanks{
$\dagger$Work done at OMRON SINIC X Corp. as part of an internship.
$^{1}$S. Katayama is with the Department of System Science, Graduate School of Informatics, Kyoto University, Kyoto, Japan
        {\tt\small katayama@sys.i.kyoto-u.ac.jp}, 
        Tatsunori Taniai and Kazutoshi Tanaka are with OMRON SINIC X Corporation, Hongo 5-24-5, Bunkyoku, Tokyo 113-0033, Japan 
        {\tt\small \{tatsunori.taniai, kazutoshi.tanaka\}@sinicx.com}}%
}

\ifarxiv
\usepackage{fancyhdr}
\usepackage{url}

\fi

\begin{document}

\ifarxiv
\twocolumn[
\noindent
© 2022 IEEE. Personal use of this material is permitted. Permission from IEEE must be obtained for all other uses, in any current or future media, including reprinting/republishing this material for advertising or promotional purposes, creating new collective works, for resale or redistribution to servers or lists, or reuse of any copyrighted component of this work in other works.\\

\noindent
\textbf{Published article:}\\
S. Katayama, T. Taniai, and K. Tanaka, ``Quasistatic contact-rich manipulation via linear complementarity quadratic programming,'' 2022 IEEE/RSJ International Conference on Intelligent Robots and Systems (IROS 2022), 2022, pp.~203–210.
]
\thispagestyle{empty}
\pagenumbering{gobble}
\clearpage
\fi
\maketitle
\thispagestyle{empty}
\ifarxiv
\else
\pagestyle{empty}
\fi

\begin{abstract}
Contact-rich manipulation is challenging due to dynamically-changing physical constraints by the contact mode changes undergone during manipulation.
This paper proposes a versatile local planning and control framework for contact-rich manipulation that determines the continuous control action under variable contact modes online.
We model the physical characteristics of contact-rich manipulation by quasistatic dynamics and complementarity constraints.
We then propose a linear complementarity quadratic program (LCQP) to efficiently determine the control action that implicitly includes the decisions on the contact modes under these constraints.
In the LCQP, we relax the complementarity constraints to alleviate ill-conditioned problems that are typically caused by measure noises or model miss-matches.
We conduct dynamical simulations on a 3D physical simulator and demonstrate that the proposed method can achieve various contact-rich manipulation tasks by determining the control action including the contact modes in real-time. 
\ifarxiv
Code is available at \url{https://omron-sinicx.github.io/lcqp/}.
\fi

\end{abstract}

\section{Introduction} 
\begin{figure}
\centering
\includegraphics[scale=0.3]{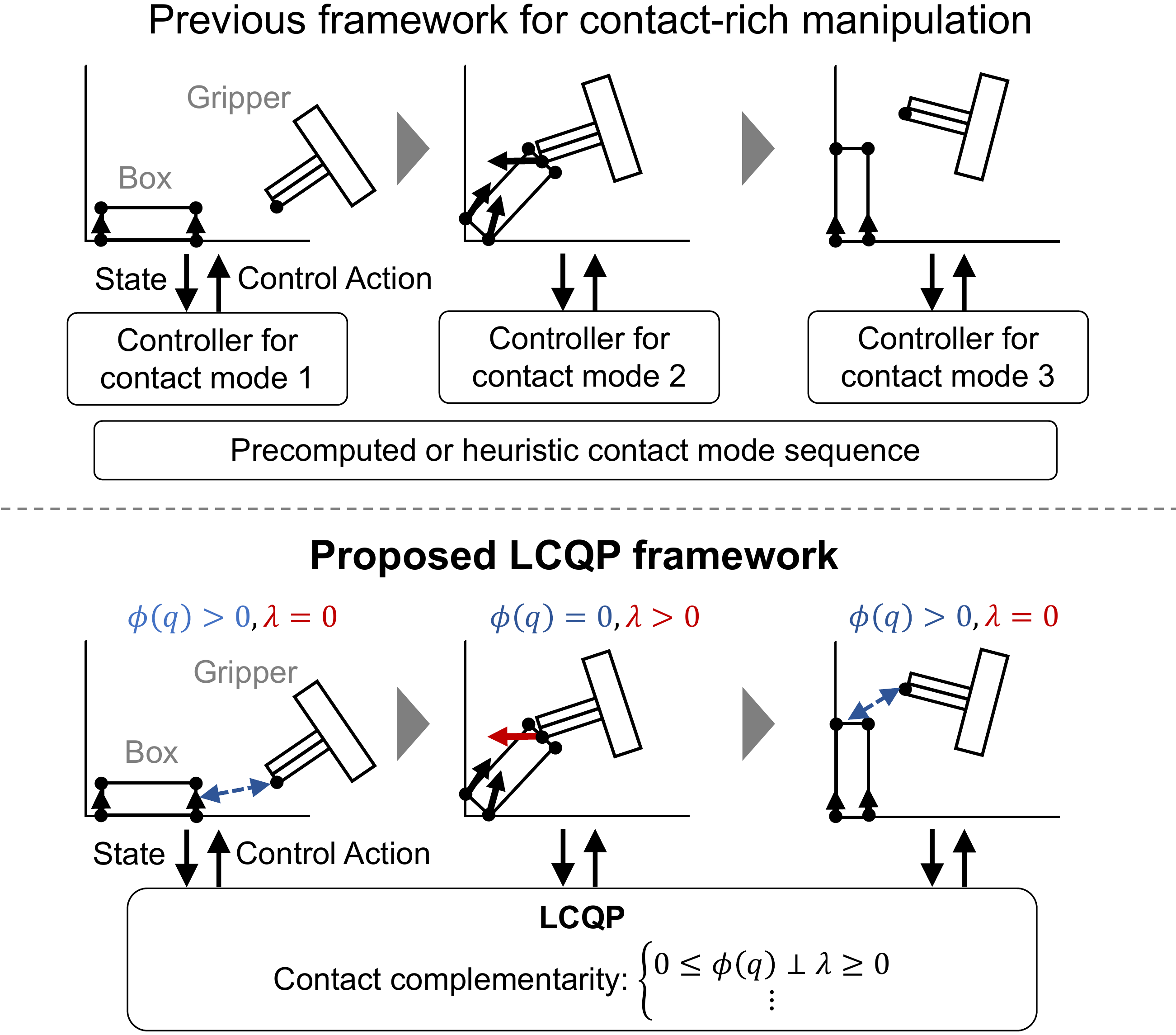}
\caption{The proposed linear complementarity quadratic programming (LCQP) framework for contact-rich manipulation. The upper figure shows an existing framework for contact-rich manipulation that requires a precomputed or heuristic contact mode sequence. The lower figure shows the proposed LCQP framework, which plans the control action and simultaneously infers the contact modes online in a principled fashion using the contact-force complementarity. 
}
\label{fig:overview}
\end{figure}

Leveraging contacts is a key feature toward human-like versatile and dexterous robotic manipulation.
For example, humans can often perform manipulation even without grasping, by utilizing the contacts between  objects and the environment. Recent studies attempt to achieve such manipulation with robots~\cite{bib:sharedGraspControl, bib:CITO-PWA, bib:quasidynamicMIP, bib:contactModeSampling2D, bib:hierarchicalManipulation}.
Fig.~\ref{fig:overview} shows an example of such manipulation, in which a gripper pivots and stands up a box by utilizing contacts between the box and the environment without grasping. 
We refer to such manipulation leveraging various contacts as contact-rich manipulation in this paper. 

Contact-rich manipulation involves many possible discrete states (contact modes) as well as the continuous states (position, velocity, and force) depending on each other. Such dynamical systems, called hybrid systems~\cite{bib:hybridDynamicalSystemsBook}, make the existing planning and control approaches complicated \cite{bib:hybridMPCOverview}.
This problem is particularly difficult in a dynamic environment, in which the precomputation results of the planning and control laws are difficult to deploy accurately and safely on real robots.
Therefore, real-time decision making including the contact modes is necessary for manipulation problems in a dynamic environment. 
Meanwhile, hybrid systems may be controlled via heuristic rules (e.g., \textit{if-then-else}) to switch between mode-specific controllers (Fig.~\ref{fig:overview}, top). However, it is preferred to avoid such rule-based methods because they lack versatility and potentially result in unreasonable, conservative, or infeasible solutions.

Generally, gradient-based planning methods, such as trajectory optimization (TO) and model predictive control (MPC), can systematically incorporate the physical reasoning about dynamical systems as constraints and are widely used for manipulation planning when decision-making on contact modes is unnecessary~\cite{bib:TOManipulation, bib:chomp, bib:trajopt, bib:MPCManipulation}.
However, once the contact modes are included in the decision variables,  application of gradient-based methods becomes difficult because the gradients are not defined at the mode switching instants (e.g., when making and breaking contacts).
For this issue,  existing studies on contact-rich planning have proposed to use mixed integer programming (MIP) \cite{bib:MIPMPCManipulation}, contact implicit trajectory optimization (CITO) \cite{bib:CITOOriginal, bib:CITO, bib:CITO-PWA}, or sampling-based planning \cite{bib:hierarchicalManipulation}.
However, these techniques are limited to offline planning due to their heavy computational complexities.
Similar computational load issues exist in stabilizing controllers for contact-rich manipulation tasks~\cite{bib:CITO-PWA, bib:complementaritySOS}.

In this study, we aim to develop a planning method for contact-rich manipulation that allows real-time execution with online reasoning about contact modes.
Our key idea, highlighted in Fig.~\ref{fig:overview}, is to utilize complementarity constraints to formulate the physical characteristics of the contacts in a principle fashion. The complementarity constraint on two scalar variables, $a$ and $b$, is expressed as
\begin{equation}
0 \leq a \perp b \geq 0,
\end{equation}
denoting that $a,b$ satisfy $a,b\geq 0$ and $ab = 0$. In Fig.~\ref{fig:overview}, the distance between the gripper and the box, $\phi(q)$, and the contact force between them, $\lambda$, are complementary with each other during manipulation, as a physical constraint. With complementarity constraints, we can express discrete contact statuses of a hybrid system without explicitly enumerating them as in MIP, thereby reducing computational burdens.
The complementarity constraint is used in  physical simulations \cite{bib:timeStepping, bib:quasistaticTimeStepping, bib:quasistaticManipulation1, bib:quasistaticManipulation2, bib:quasistaticManipulation3} and has been applied for planning \cite{bib:CITO, bib:CITO-PWA, bib:CITODynamicManipulation}.
While the existing planners using complementarity constraints still involve large computational complexities, we  formulate the planning and control problem as a tractable quadratic program (QP) with linear complementarity constraints, called linear complementarity quadratic program (LCQP) \cite{bib:LCQPSolver}.
The proposed method, named LCQP controller, is derived via a quasistatic model, which can reduce the problem complexity while maintaining the ability to handle contact-rich manipulation~\cite{bib:quasistaticManipulation1, bib:quasistaticManipulation2, bib:quasistaticManipulation3, bib:contactModeSampling2D, bib:HybridForceVelocity}.
Furthermore, to increase the robustness of the proposed method in practical situations where the constraints are violated by the model mismatches or sensor noises, we adopt relaxed  complementarity constraints.

The contributions of this paper are summarized as follows.
\begin{enumerate}
    \item We formulate the local planning and control of contact-rich manipulation problems as an LCQP, which is affordable for online execution. 
    \item We address the ill-conditioning of the LCQP in practical situations by relaxing the complementarity constraints. 
    \item We conduct dynamical simulations on a 3D physical simulator, PyBullet \cite{bib:pybullet}, and demonstrate that the proposed method can determine  continuous control action while inferring the contact modes online. 
\end{enumerate}

The remainder of this paper is organized as follows.
Section \ref{section:relatedWork} introduces related work on the contact-rich manipulation planning and control.
Section \ref{section:quasistaticFormulation} models the quasistatic manipulation problem using the complementarity constraints.
Section \ref{section:LCQPFormulation} proposes the LCQP for the manipulation control problem.
Section \ref{section:Simulation} demonstrates the effectiveness of the proposed LCQP controller through simulations on the 3D simulator PyBullet. 
Finally, a brief summary and mention of future work are presented in Section \ref{section:Conclusion}.

\section{Related Work}\label{section:relatedWork}
\subsection{Contact-rich manipulation planning}\label{section:relatedWorkPlanning}
Contact-rich manipulation planning involves non-differentiable constraints due to contacts.
The following three approaches have been studies for this problem.

\subsubsection{Mixed integer programming}
The most straightforward approach is to incorporate the contact modes explicitly as integer variables in a MIP~\cite{bib:quasidynamicMIP}.
However, the computational cost of the MIP grows exponentially with respect to the length of the planning time interval and possible contact modes.
Thus, approximations via learned mode schedules are necessary for online execution of MPC \cite{bib:MIPMPCManipulation}.
While these MIP-based methods tend to be burdened by explicit enumeration of the contact modes, the proposed framework can avoid such enumeration via contact complementarity. 

\subsubsection{Contact implicit trajectory optimization}
The second approach is to model the contacts as complementarity constraints~\cite{bib:timeStepping, bib:quasistaticTimeStepping} in the trajectory optimization framework. The planning problem is then formulated as a mathematical program with complementarity constraints (MPCC), which is also termed as CITO \cite{bib:CITOOriginal, bib:CITO, bib:CITO-PWA}.
However, the MPCC is inherently ill-conditioned and the numerical solvers often take a long computational time or fail to find solutions because it lacks the constraint qualifications~\cite{bib:mpcc}. 
Moreover, numerical alleviation of the problem often leads to suspicious stationary points \cite{bib:MPCCDifficulty}.
While the above studies~\cite{bib:timeStepping, bib:quasistaticTimeStepping} consider the nonlinear optimization problem with nonlinear complementarity constraints, the present method considers a tractable QP with linear complementarity constraints by simplifying the problem formulation. 

\subsubsection{Sampling-based planning}
The third approach first determines a contact-mode trajectory via sampling-based methods~\cite{bib:motionConesampling, bib:hierarchicalManipulation, bib:contactModeSampling2D, bib:contactModeSampling3D} and then plans or executes the control actions of the robot while following the modes. 
Although this disjoint optimization of contact modes and actions can improve the computational time and robustness over plain CITO for particular problems~\cite{bib:trajOptAndTreeSearch}, they are still limited to the off-line computation.

\subsection{Contact-rich manipulation control}\label{section:relatedWorkControl}
Because control of contact-rich manipulation must consider contact forces between objects and the environment along with actuation forces, its implementation is non-trivial even if optimized contact modes and trajectories are available. This is especially the case for non-prehensile manipulation~\cite{bib:nonprehensileManipulation, bib:nonprehensileManipulationSurvey}, in which a robot largely relies on such indirect forces to perform manipulation without grasping (i.e., without form closure or force closure~\cite{bib:forceClosure}). For example, shared grasping~\cite{bib:sharedGraspControl} performs certain manipulation tasks without grasping objects by robot hand. A successful approach for the shared grasping is hybrid force-velocity control~\cite{bib:HybridForceVelocity, bib:forceVelocityControl2} based on quasistatic manipulation models.

Our LCQP controller also employs a quasistatic model, and jointly optimizes actuation forces, environmental contact forces, and other geometric configurations of the robot and objects to be taken at a next time step. We use these optimized estimates to implement a simple PD-based joint controller, as we will explain in Sections~\ref{section:jointController} and \ref{subsection:details}.

For more advanced contact-rich manipulation, some studies further assume acceleration-level dynamics models \cite{bib:CITO-PWA, bib:quasidynamicMIP}. However, these models generally require heavy offline computations such as CITO~\cite{bib:CITO-PWA} and MIP~\cite{bib:quasidynamicMIP} to construct controllers as well as trajectories. As discussed in Section I, such offline computations lack versatility when applied in a dynamic environment.

In this paper, we focus on online planning under variable contact modes rather than achieving complicated motion control via acceleration-level dynamics. Extending the proposed framework to deal with these control issues is future work.

\section{Complementarity Formulation of Quasistatic Robotic Manipulation}\label{section:quasistaticFormulation}
Before presenting the proposed LCQP, let us introduce the physical constraints of contact-rich manipulation problems, namely, a quasistatic dynamical model and the contact  complementarity, that are used in the method. The former describes the time evolution law of manipulation processes that the method needs to follow. The latter further enforces the manipulation planning to meet the physical characteristics of the contact forces.
For notational simplicity, we here focus on planar motion problems. However, it is easy to extend the present method to a three-dimensional formulation by approximating the friction cone using polyhedral~\cite{bib:quasistaticTimeStepping}.

\subsection{Quasistatic assumption}

\begin{figure}[tb]
\centering
\includegraphics[scale=0.4]{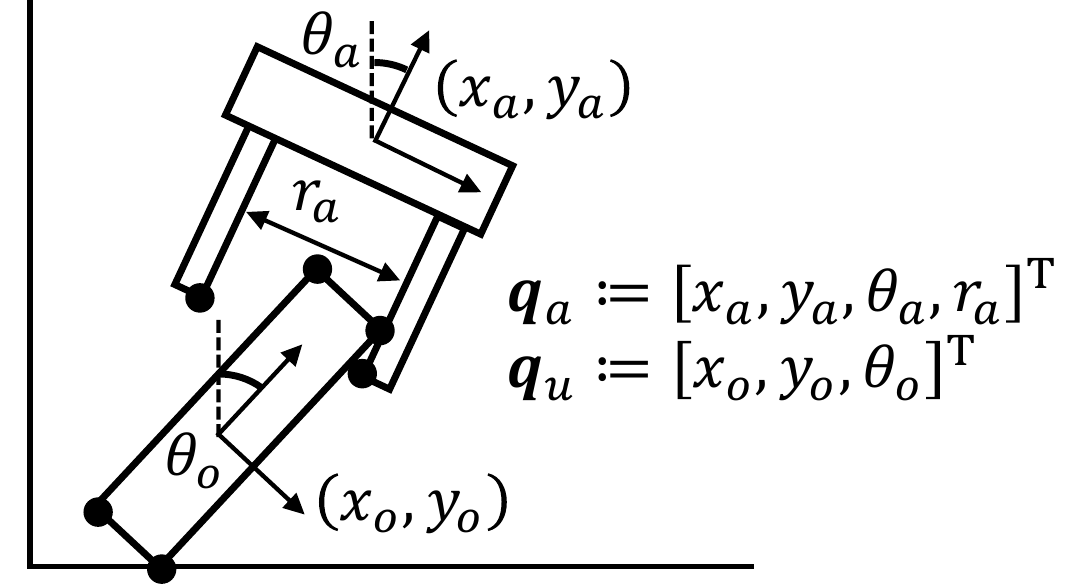}
\caption{An example of quasistatic formulation where the total dynamical system is composed of a parallel gripper (actuated system) and a box (underactuated system). 
Black dots represent candidates of the contact points. 
The contact complementarity constraints between each contact point candidate and nearby surfaces (straight lines) are considered.}
\label{fig:LCP_example}
\end{figure}

We model the actuated systems (i.e., robots or grippers) and the underactuated systems (i.e., objects to be manipulated) as a quasistatic dynamical model, in which the velocity and acceleration are sufficiently small.
This model can make the problem formulation simple while being sufficiently accurate for certain contact-rich problems \cite{bib:quasistaticTimeStepping, bib:quasistaticManipulation1, bib:quasistaticManipulation2, bib:quasistaticManipulation3}.
Let $N_a$ be the total degree of freedom (DOF) of the actuated systems and $N_u$ be that of the underactuated systems.
We denote the stack of the configurations (generalized coordinates) of the actuated and underactuated systems as $\bm{q}_a \in \mathbb{R}^{N_a}$ and $\bm{q}_u \in \mathbb{R}^{N_u}$, respectively.
Fig.~\ref{fig:LCP_example} depicts an example of this formulation with a parallel gripper as the actuated system and a box as the underactuated system.
The state of the total dynamical system $\bm{q}$ is composed of the configurations of all the systems, i.e., $\bm{q} = [\bm{q}_a ^{\rm T} \;\; \bm{q}_u^{\rm T} ]^{\rm T}$.
With the generalized velocity $\bm{v} = [\bm{v}_a ^{\rm T} \;\; \bm{v}_u ^{\rm T} ]^{\rm T}$, the discrete-time expansion of the quasistatic model is expressed as
\begin{equation}\label{eq:stateEquation}
    \bm{q}_{t+1} = \bm{q}_t + h \bm{v}_t,
\end{equation}
where the subscripts $t$ and $t+1$ denote the time steps of $\bm{q}$ and $\bm{v}$, and $h > 0$ denotes the time elapsed from $t$ to $t+1$. Here, $h$ is assumed to be sufficiently small.

Next, we introduce the force and moment balance on objects to incorporate the Newtonian dynamics under the quasistatic assumption into the proposed LCQP. To do so, we first formulate the contact force based on the contact between a point and a surface.
As illustrated in Fig.~\ref{fig:LCP_example}, we consider corners of the objects and those of the parallel gripper as possible contact points (black dots) and evaluate their contacts with nearby surfaces (straight lines). 
We call each of such point-surface pairs a ``contact candidate.''
Each contact candidate $i$ is associated with a contact force $\bm{\lambda}_{t, i} = [\lambda_{f_i}, \lambda_{n_i}] ^{\rm T} \in \mathbb{R}^2$ at each time $t$, where $\lambda_{f_i},\lambda_{n_i} \in \mathbb{R}$ represent its tangential contact force and normal contact force, respectively.
Note that other kinds of contacts such as rolling contacts can also be formulated as complementarity constraints.
As such, we can consider other types of shapes than boxes, e.g., disks.

To derive the force and moment balance for each underactuated object $o$, we further introduce $C_o$ to denote a set of contact candidates associated with an object $o$. Each candidate $i \in C_o$ (i.e., a point and plane pair) is associated with the rotation matrix of the contact plane, $\bm{R}_i (\bm{q}_t) \in \mathbb{R}^{2 \times 2}$, as well as the relative position of the contact point from the center of mass of the object $o$ expressed in the world coordinates, $[r_{o, i, x} (\bm{q}_t) \;\, r_{o, i, y} (\bm{q}_t)]^{\rm T}$.
Under the quasistatic assumption (i.e., the velocity and acceleration being sufficiently small), the Newton's second law for each underactuated object $o$ at time $t$ reduces to the force and moment balance:
\begin{equation}\label{eq:forceBalance}
    \bm{b}_o (\bm{q}_t, \left\{ \bm{\lambda}_{t, i} \right\}_{i \in C_o} ) := \sum_{i \in C_o} \begin{bmatrix} \bm{R}_i (\bm{q}_t) \bm{\lambda}_{t, i} \\ \tilde{\bm{r}}_{o, i} ^{\rm T} (\bm{q}_t) \bm{R}_i (\bm{q}_t) \bm{\lambda}_{t, i} \end{bmatrix} + \bm{F}_o = \bm{0},
\end{equation}
where $\tilde{\bm{r}}_{o, i} (\bm{q}_t) := [- r_{o, i, y} (\bm{q}_t) \;\, r_{o, i, x} (\bm{q}_t)] ^{\rm T} \in \mathbb{R}^2$.
Here, Eq.~(\ref{eq:forceBalance}) in its first and second dimensions represents the total force, expressed in the world coordinates, that are applied on the object $o$. Likewise, Eq.~(\ref{eq:forceBalance}) in the third dimension represents the total moment. The forces and moments in the bracket represent the contact forces on the object, while $\bm{F}_o \in \mathbb{R}^3$ represents the sums of the temporally constant forces and moments such as the gravity.

\subsection{Contact complementarity}
\label{section:contactComplementarity}
\begin{figure}[tb]
\centering
\includegraphics[scale=0.55]{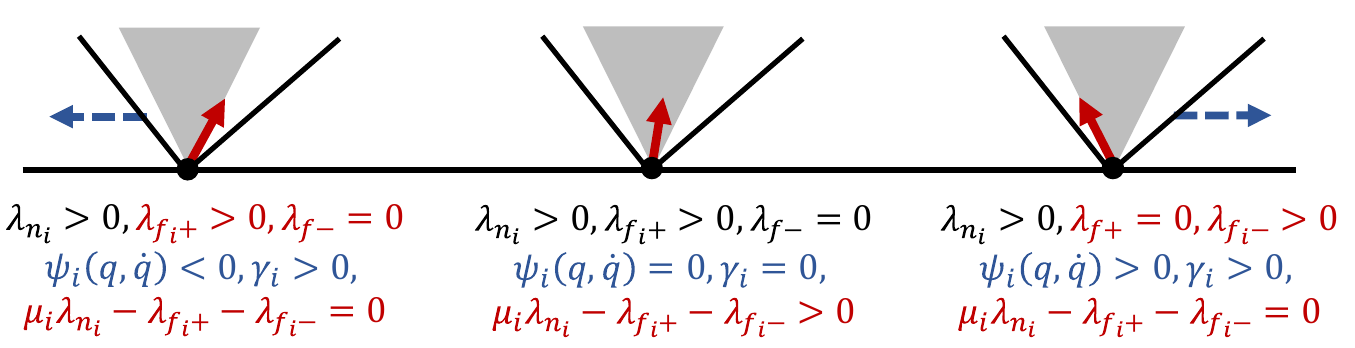}
\caption{Illustration of the contact complementarity constraints associated with the tangential direction of the contact surface. The contact between a corner of a box (black dots) and the ground is shown. Blue dashed lines are tangential velocities. Red solid lines are contact forces. Gray regions show the friction cones.}
\label{fig:lcp_tan_illustration}
\end{figure}

We utilize the contact complementarity to describe the characteristics of the contact force $\bm{\lambda}_{t,i}$: for each contact candidate $i$,  (a)  the normal force $\lambda_{f_i}$ is positive only when the distance between the contact point and surface is zero (see Fig.~\ref{fig:overview}), and (b) the contact force $\bm{\lambda}_{t,i}$ must lie within a friction cone defined by  Coulomb's friction coefficient (see Fig.~\ref{fig:lcp_tan_illustration}). 
For (a), we introduce $\phi_i (\bm{q}_t) \in \mathbb{R}$ as the distance between the contact point and contact surface of the contact candidate $i$ at time $t$.
For (b), we also introduce $\psi_i (\bm{q}_t, \bm{v}_t) \in \mathbb{R}$ as the tangential velocity of the contact point expressed in the local coordinate of the contact surface at time $t$.
Furthermore, in order to express the characteristic (b) as linear complementarity constraints~\cite{bib:timeStepping, bib:quasistaticTimeStepping}, we decompose the tangential contact force $\lambda_{f_i}$ into the two non-negative tangential contact forces in the positive and negative directions, $\lambda_{f_i +}, \lambda_{f_i -} \in \mathbb{R}_+$, as $\lambda_{f_i} = \lambda_{f_i +} - \lambda_{f_i -}$.

The contact complementarity constraints for the $i$-th contact candidate are then expressed as \cite{bib:timeStepping, bib:quasistaticTimeStepping}:
\begin{subequations}\label{subeq:contactComplementatiry}
\begin{equation}\label{eq:normalContactComplementarity}
    0 \leq \lambda_{n_i} \perp \phi_i (\bm{q}_{t}) \geq 0,
\end{equation}
\begin{equation}\label{eq:tangentialContactComplementarity1}
    0 \leq \lambda_{f_i +} \perp \gamma_i + \psi_i (\bm{q}_{t}, \bm{v}_{t}) \geq 0,
\end{equation}
\begin{equation}\label{eq:tangentialContactComplementarity2}
    0 \leq \lambda_{f_i -} \perp \gamma_i - \psi_i (\bm{q}_{t}, \bm{v}_{t}) \geq 0,
\end{equation}
\begin{equation}\label{eq:tangentialVelocityComplementarity}
    0 \leq \gamma_{i} \perp \mu_i \lambda_{n_i} - \lambda_{f_i +} - \lambda_{f_i -} \geq 0,
\end{equation}
\end{subequations}
where $\mu_i > 0$ is the friction coefficient of the $i$-th contact candidate and $\gamma_i \in \mathbb{R}$ is an auxiliary variable.
$\gamma_i$ is treated as an optimization varible in the LCQP and typically takes the magnitude of the tangential contact velocity $|\psi_i (\bm{q}_{t}, \bm{v}_{t})|$. 
Here, Eq.~(\ref{eq:normalContactComplementarity}) corresponds to the above characteristic (a), while Eqs.~(\ref{eq:tangentialContactComplementarity1})--(\ref{eq:tangentialVelocityComplementarity}) correspond to (b). 
Particularly, Eqs.~(\ref{eq:tangentialContactComplementarity1}) and (\ref{eq:tangentialContactComplementarity2}) ensure that either $\lambda_{f+}$ or $\lambda_{f-}$ is zero when $\psi_i (\bm{q}_t, \bm{v}_t)$ is non-zero. This is because either $\gamma_i + \psi_i (\bm{q}_t, \bm{v}_t)$ or $\gamma_i - \psi_i (\bm{q}_t, \bm{v}_t)$ is non-zero when $\psi_i (\bm{q}_t, \bm{v}_t)$ is non-zero.
This relation also derives a property of friction forces that the tangential velocity $\psi_i (\bm{q}_t, \bm{v}_t)$ is opposite in sign to the force $\lambda_{f_i} = \lambda_{f_i +} - \lambda_{f_i -}$.
Finally, Eq.~(\ref{eq:tangentialVelocityComplementarity}) ensures that the contact force $\bm{\lambda}_{t,i}$ sticks to the friction cone if the tangential velocity is non-zero (see Fig.~\ref{fig:lcp_tan_illustration}, left and right).

Note that when the proposed method is applied to the case of 3D motions,  a pair of (\ref{eq:tangentialContactComplementarity1}) and (\ref{eq:tangentialContactComplementarity2}) is introduced for each diagonal of the polyhedral cone~\cite{bib:quasistaticTimeStepping}.

\section{Quasistatic Manipulation Control via Linear Complementarity Quadratic Programs}\label{section:LCQPFormulation}
In this section, we formulate the LCQP controller. 
Our goal with the LCQP controller is \emph{finding the position and force commands for the actuated systems to achieve given tasks while satisfying the physical constraints derived in the previous section}. To this end, let us define $\bm{\lambda}_t$ as the stack of $[\lambda_{f_i + }, \lambda_{f_i - }, \lambda_{n_i}] ^{\rm T} \in \mathbb{R}^3$ for all contact candidates~$i$, and introduce an objective function to be minimized with respect to $\bm{q}_{t+1}$ and $\bm{\lambda}_{t+1}$ satisfying the physical constrains (\ref{eq:stateEquation})--(\ref{eq:tangentialVelocityComplementarity}):
\begin{align}\label{eq:costOriginal}
    & \frac{1}{2} (\bm{g} (\bm{q}_{t+1}) - \bm{g}_{\rm ref})^{\rm T} \bm{Q}_{g} (\bm{g} (\bm{q}_{t+1}) - \bm{g}_{\rm ref}) \notag \\
    & + \frac{1}{2} \bm{v}_t ^{\rm T} \bm{Q}_{v} \bm{v}_t + \frac{1}{2} \bm{\lambda}_{t+1} ^{\rm T} \bm{Q}_{f} \bm{\lambda}_{t+1},
\end{align}
where $\bm{g} (\bm{q})$ is a general differentiable function mapping a state $\bm{q}$ into a task space such as positions and orientations of objects, $\bm{g}_{\rm ref}$ denotes their goal values, and $\bm{Q}_{g}$, $\bm{Q}_v$, and $\bm{Q}_f$ are given positive definite weight matrices. Here,
$\bm{Q}_{g}$ represents priorities of tasks, and $\bm{Q}_v$ and $\bm{Q}_f$ are regularization terms. 

In the LCQP controller described below, we optimize this constrained nonlinear objective function analogously to gradient descent.  By making a descent step for Eq.~(\ref{eq:costOriginal}), we plan and execute a control action in a manipulation process. Each descent step is computed by solving a linearized objective function, resulting in the desired position and force commands, $\bm{q}_{t+1}$ and $\bm{\lambda}_{t+1}$, for the next time step. We expect that iterating between solving a linearized objective and executing a command through a manipulation process will eventually reach a goal state $\bm{q}^*$ as the minimizer of Eq.~(\ref{eq:costOriginal}).

\subsection{Problem linearization}
To derive a tractable problem to be solved at each time step, we define a QP with respect to $\Delta \bm{q}_t$ and $\Delta \bm{\lambda}_t$ that are defined from $\bm{q}_{t+1} = \bm{q}_{t} + \Delta \bm{q}_t$ and $\bm{\lambda}_{t+1} = \bm{\lambda}_t + \Delta \bm{\lambda}_t$.
When $\Delta \bm{q}_t$ and $\Delta \bm{\lambda}_t$ are sufficiently small,
the objective function (\ref{eq:costOriginal}) is  approximated by a quadratic function of $\Delta \bm{q}_t$ and $\Delta \bm{\lambda}_t$:
\begin{align}\label{eq:QPcost}
    & \frac{1}{2} (\bm{g}_t + \bm{G}_t \Delta \bm{q}_{t} - \bm{g}_{\rm ref})^{\rm T} \bm{Q}_{g} (\bm{g}_t + \bm{G}_t \Delta \bm{q}_{t} - \bm{g}_{\rm ref}) \notag \\
    & + \frac{1}{2} \Delta \bm{q}_{t} ^{\rm T} \bm{Q}_{\Delta \bm{q}} \Delta \bm{q}_{t} + \frac{1}{2} (\bm{\lambda}_{t} + \Delta \bm{\lambda}_{t}) ^{\rm T} \bm{Q}_{f} (\bm{\lambda}_{t} + \Delta \bm{\lambda}_t),
\end{align}
where we define $\bm{g}_t := \bm{g} (\bm{q}_t)$, $\bm{G}_t := \frac{\partial \bm{g}}{\partial \bm{q}} \bm{g} (\bm{q}_t)$, and $\bm{Q}_{\Delta \bm{q}} := \frac{1}{h^2} \bm{Q}_v$, respectively. Here, the constraint of the quasistatic time evolution (\ref{eq:stateEquation}), i.e., $\Delta \bm{q}_t = h \bm{v}_t$, is incorporated into Eq.~(\ref{eq:QPcost}). The other physical constraints (\ref{eq:forceBalance})--(\ref{eq:tangentialVelocityComplementarity}) are further linearized below to derive a tractable LCQP formulation.

\subsubsection{Contact complementarity constraints}

The contact complementarity constriants (\ref{eq:normalContactComplementarity})--(\ref{eq:tangentialVelocityComplementarity}) with respect to $q_{t+1}$, $\lambda_{n_i}$, $\lambda_{f_i +}$, and $\lambda_{f_i -}$ are approximated by linear complementrity constraints with respect to $\Delta \bm{q}_t$, $\Delta \lambda_{n_i}$, $\Delta \lambda_{f_i +}$, and $\Delta \lambda_{f_i -}$ as:
\begin{subequations}\label{eq:LCP}
\begin{equation}\label{eq:normalLCP}
    0 \leq \lambda_{n_i} + \Delta \lambda_{n_i} \perp \phi_{i} (\bm{q}_{t}) + \bm{\Phi}_{i, \bm{q}} (\bm{q}_t) \Delta \bm{q}_t \geq 0,
\end{equation}
\begin{equation}\label{eq:tangentialLCP1}
    0 \leq \lambda_{f_i +} + \Delta \lambda_{f_i +} 
    \perp \gamma_i + \psi_{i} (\bm{q}_t, \bm{v}_t) + \bm{\Psi}_{i, \bm{v}} (\bm{q}_t, \bm{v}_t) \Delta \bm{q}_t \geq 0,
\end{equation}
\begin{equation}\label{eq:tangentialLCP2}
    0 \leq \lambda_{f_i -} + \Delta \lambda_{f_i +}  
    \perp \gamma_i - \psi_{i} (\bm{q}_t, \bm{v}_t) - \bm{\Psi}_{i, \bm{v}} (\bm{q}_t, \bm{v}_t) \Delta \bm{q}_t \geq 0,
\end{equation}
and
\begin{align}\label{eq:tangentialVelocityLCP}
    0 \leq \gamma_{i} \perp & \mu_i (\lambda_{n_i} + \Delta \lambda_{n_i}) \notag \\ 
    & - \lambda_{f_i +} - \Delta \lambda_{f_i +} - \lambda_{f_i -} - \Delta \lambda_{f_i -} \geq 0,
\end{align}
\end{subequations}
where  $\bm{\Phi}_{i, \bm{q}} (\bm{q}) := \frac{\partial \bm{\phi}}{\partial \bm{q}} (\bm{q})$ and $\bm{\Psi}_{i, \bm{v}} (\bm{q}, \bm{v}) := \frac{\partial \bm{\psi}}{\partial \bm{v}} (\bm{q}, \bm{v})$.
Note that when solving the LCQP, $\bm{v}_t = \frac{\Delta \bm{q}_t}{h} = \bm{0}$ is substituted in Eqs. (\ref{eq:normalLCP})--(\ref{eq:tangentialVelocityLCP}) due to the quasistatic assumption.

\subsubsection{Force and moment balance}
The constraint of the force and moment balance (\ref{eq:forceBalance}) is linearized as 
\begin{align}\label{eq:forceBalanceLinearized}
    & \bm{b} (\bm{q}_t, \left\{ \bm{\lambda}_{t, j} \right\}) 
    + \frac{\partial \bm{b}}{\partial \bm{q}_t} (\bm{q}_t, \left\{\bm{\lambda}_{t, i} \right\}_{i \in C_0}) \Delta \bm{q}_t \notag \\ 
    & + \sum_{i \in C_0} \frac{\partial \bm{b}}{\partial \bm{\lambda}_{t, i}} (\bm{q}_t, \left\{\bm{\lambda}_{t, i}\right\}_{i \in C_0}) \Delta \bm{\lambda}_{t, i} = \bm{0}
\end{align}
and imposed in the LCQP as an equality constraint.

\subsection{Relaxing contact complementarity constraints}

The complementarity constraints (\ref{eq:normalLCP})--(\ref{eq:tangentialLCP2}) can cause numerical issues.
First, solutions satisfying the complementarity constraints (\ref{eq:normalLCP})--(\ref{eq:tangentialLCP2}) may not exist because the measured states can have unrealistic penetrations of objects due to sensor noises or model miss-matches (e.g., modeling errors in the shapes of objects or deformations of the objects). By this issue, the solver may fail to find a solution.
Second, the complementarity constraints (\ref{eq:normalLCP})--(\ref{eq:tangentialVelocityLCP}) often prevent the actuated objects from contacting with the objects in the real system.
For example, even when the LCQP solver computes a non-zero normal contact force command, the real position of the contact point may slightly deviate from the surface.
To solve these issues, we need to allow small penetrations of objects in numerical computations to ensure physical contacts of objects in the real system. 

To this end, we relax the complementarity constraints (\ref{eq:normalLCP})--(\ref{eq:tangentialVelocityLCP}) by introducing a slack variable $s_i$ as
\begin{subequations}\label{eq:relaxedLCP}
\begin{equation}\label{eq:relaxedNormalLCP}
    0 \leq \lambda_{n_i} + \Delta \lambda_{n_i} + s_i \perp \phi_{i} (\bm{q}_{t}) + \bm{\Phi}_{\bm{q}, i} (\bm{q}_t) \Delta \bm{q}_t + s_i \geq 0,
\end{equation}
\begin{align}\label{eq:relaxedTangentialLCP1}
    0 \leq \; & \lambda_{f_i +} + \Delta \lambda_{f_i +} + s_i \notag \\ 
    & \perp \gamma_i + \psi_{i} (\bm{q}_t, \bm{v}_t) + \bm{\Psi}_{\bm{v}, i} (\bm{q}_t, \bm{v}_t) \Delta \bm{q}_t + s_i \geq 0,
\end{align}
\begin{align}\label{eq:relaxedTangentialLCP2}
    0 \leq \; & \lambda_{f_i -} + \Delta \lambda_{f_i +} + s_i \notag \\ 
    & \perp \gamma_i - \psi_{i} (\bm{q}_t, \bm{v}_t) - \bm{\Psi}_{\bm{v}, i} (\bm{q}_t, \bm{v}_t) \Delta \bm{q}_t + s_i \geq 0,
\end{align}
and
\begin{align}\label{eq:relaxedTangentialVelocityLCP}
    0 \leq \gamma_{i} + s_i \perp \; & \mu_i (\lambda_{n_i} + \Delta \lambda_{n_i}) - \lambda_{f_i +} - \Delta \lambda_{f_i +} \notag \\ 
    & - \lambda_{f_i -} - \Delta \lambda_{f_i -} + s_i \geq 0.
\end{align}
\end{subequations}
We then add  $p_i s_i ^2$ to the objective function (\ref{eq:QPcost}),  where $p_i$ is a sufficiently large positive penalty parameter.

Note that in the context of quasistatic time-stepping simulations \cite{bib:quasistaticManipulation1, bib:quasistaticManipulation2, bib:quasistaticManipulation3}, similar problems associated with numerical errors are reported. These errors can cause numerical ill-conditioning for our LCQP, even in the absence of the aforementioned issues such as model inaccuracy. The proposed relaxation is effective to alleviate this issue too.

\subsection{Linear complementarity quadratic program (LCQP)}
\subsubsection{Summary of the LCQP}
We herein summarize the LCQP formulation for the quasistatic contact-rich manipulation control.
Let $\bm{s}_t$ be the stack of $\gamma_i$ and $s_i$ for all contact candidates $i$.
The LCQP is given as
\begin{align}\label{eq:LCQP}
    \min_{\Delta \bm{q}_t, \bm{\lambda}_{t+1}, \bm{s}_t} & \frac{1}{2} (\bm{g}_t + \bm{G}_t \Delta \bm{q}_t - \bm{g}_{\rm ref})^{\rm T} \bm{Q}_{g} (\bm{g}_t + \bm{G}_t \Delta \bm{q}_t - \bm{g}_{\rm ref}) \notag \\
    & + \frac{1}{2} \Delta \bm{q}_{t} ^{\rm T} \bm{Q}_{\Delta \bm{q}_{t}} \Delta \bm{q}_{t} \notag \\ 
    & + \frac{1}{2} (\bm{\lambda}_{t} + \Delta \bm{\lambda}_t) ^{\rm T} \bm{Q}_{f} (\bm{\lambda}_{t} + \Delta \bm{\lambda}_t) + \sum_{i} p_i s_i ^2 \notag \\ 
    {\rm s.t.} \;\;\; & (\ref{eq:forceBalanceLinearized})-(\ref{eq:relaxedTangentialVelocityLCP}).
\end{align}
It is worth noting that the LCQP can be solved efficiently and reliably thanks to the recent advances in solution algorithms \cite{bib:lcqpow}.
Also note that we can additionally introduce other equality and inequality constraints than the force and moment balance (\ref{eq:forceBalanceLinearized}) such as upper and lower bounds for the optimization variables (see Section \ref{subsection:details} for details).

\subsubsection{Joint controller to track the LCQP solution}\label{section:jointController}

\begin{figure}[tb]
\centering
\includegraphics[scale=0.33]{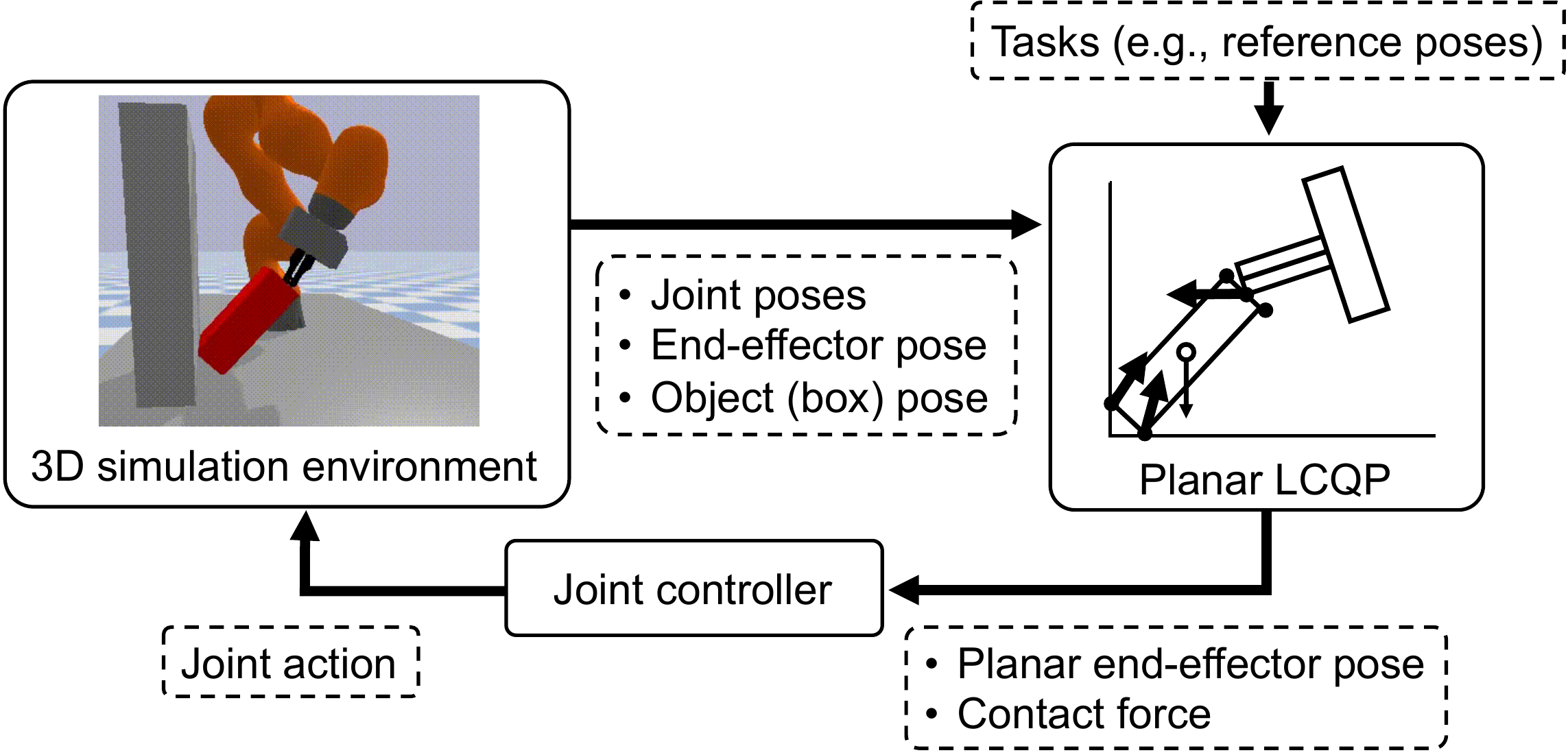}
\caption{The closed-loop control framework of the proposed method with the 3D simulation environment. Inputs and outputs of each workflow step are summarized.}
\label{fig:control_loop}
\end{figure}

By solving the LCQP in Eq.~(\ref{eq:LCQP}), we obtain $\Delta \bm{q}_t$ and $\Delta \bm{\lambda}_t$ and thus obtain the desired position and contact forces of the actuated system (e.g., a gripper) at the next time step as $\bm{q}_{t+1} = \bm{q}_t + \Delta \bm{q}_t$ and $\bm{\lambda}_{t+1} = \bm{\lambda}_t + \Delta \bm{\lambda}_t$. 
Here, we implement a joint controller of the robot manipulator composed of: 1) a feedforward torque composed by gravity compensation and 2) a PD feedback controller for the joint position and/or the end-effector pose of a robot manipulator.
In 2), we convert the force command from the LCQP solution into position command of the end-effector as
$\tilde{\bm{q}}_{t+1} = {\bm{q}}_{t+1} - \frac{1}{K} \bm{\lambda}_{t+1}$, 
where $K > 0$ is a user-defined gain.
This conversion can be expected to mitigate an issue that the robot cannot exactly execute the force command computed by LCQP in practice.
Fig.~\ref{fig:control_loop} illustrates the LCQP and joint controller with their inputs and outputs in the closed-loop control framework, which are used later in the dynamical simulation studies.

\subsubsection{Possible limiations of the proposed LCQP controller}\label{sec:limitation}
We here discuss the possible limitations of the proposed method. First, the proposed method is limited to quasistatic motions.
To achieve more dynamic motions, acceleration must be included in the LCQP formulation instead of the force and moment balance (\ref{eq:forceBalanceLinearized}) \cite{bib:quasidynamicMIP, bib:contactModeSampling3D}.
Second, the proposed method only predicts a 2D state one step ahead and can be limited to moderately simple 2D cases. 
More complicated tasks require multi-step nonlinear prediction (i.e., MPC) of 3D motions, which is challenging for online execution. 
Third, in the control framework of Fig. \ref{fig:control_loop}, the joint controller may not be able to track the LCQP solution depending on tasks. Possible options to resolve this problem include: enforcing the orthogonality between the velocity and force~\cite{bib:HybridForceVelocity, bib:forceVelocityControl2} or replacing the gripper model with a manipulator kinematics model.
Finally, the proposed method requires accurate parameters and the geometries of the system, such as the shapes, mass, and friction coefficients of the objects.

\section{Dynamical Simulations}\label{section:Simulation}
We demonstrated the effectiveness of the proposed LCQP controller by dynamical simulations on the PyBullet 3D physics simulator~\cite{bib:pybullet}, where various contact-rich manipulation tasks were conducted by the Kuka iiwa14 equipped with the Schunk WSG50 parallel gripper.
The simulations are to verify that 1) the LCQP controller can handle various contact-rich tasks, 2) an online LCQP computation is affordable, 3) the relaxation of the complementarity constraints improves the  robustness, and 4) the LCQP controller can achieve tasks even with the presence of model mismatches and measurement noises.

\subsection{Implementation details}\label{subsection:details}
First, we introduce implementation details of the LCQP controller used in the 3D dynamical simulations.

\subsubsection{Additional inequality constraints}
We added bound constraints on the configuration, velocity, and contact forces to the LCQP as
$\bm{q}_{\rm min} < \bm{q}_t + \Delta \bm{q}_t < \bm{q}_{\rm max}$,
$- \bm{v}_{\rm max} < \frac{\Delta \bm{q}_t}{h} < \bm{v}_{\rm max}$,
and $\bm{\lambda}_t + \Delta \bm{\lambda}_{t} < \bm{\lambda}_{\rm max}$, where $\bm{q}_{\rm min}, \bm{q}_{\rm max}, \bm{v}_{\rm max}$ and $\bar{\bm{\lambda}}_{i, {\rm max}}$ are constant bounds. Note that
$\bm{v}_{\rm max}$ must be small to keep the quasistatic assumption.
In addition, we impose inequality constraints to avoid unintended collisions.

\subsubsection{Joint controller}
The control loop of the LCQP controller (Fig. \ref{fig:control_loop}) is further specified as follows: 1) measure the 3D states of the gripper and target objects (cubes or cylinders), 2) construct a 2D LCQP problem by projecting the 3D states onto a 2D vertical plane, 3) solve the LCQP, 4) compute a reference 3D force and a position for the end-effector from the LCQP solution, and 5) execute the robot control. In the step 3, we solved the LCQP in Eq.~(\ref{eq:LCQP}) by using an open-source LCQP solver~\cite{bib:lcqpow} implementing an efficient solution algorithm of~\cite{bib:LCQPSolver}.

\subsection{Non-grasp pivoting of a box}\label{subsection:example1}
\subsubsection{Problem and LCQP settings}

\begin{figure}[tb]
\centering
\includegraphics[scale=0.27]{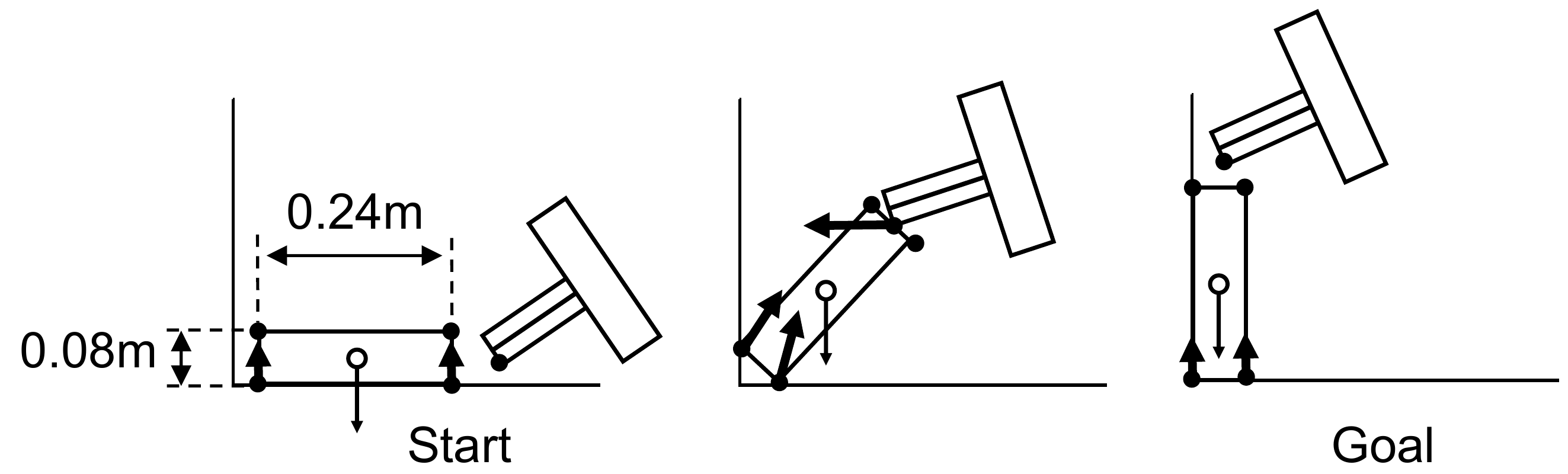}
\caption{The LCQP  configuration for pivoting a box without grasping. 
The horizontal and vertical lines are ground and wall, respectively. The black dots indicate the contact candidates. Thick arrows show the contact forces and thin arrows show the gravity forces. }
\label{fig:pivot}
\end{figure}

The first example is standing up a box by pivoting instead of grasping.
The LCQP problem setting is illustrated in Fig. \ref{fig:pivot}.
We considered the four contact points of the box (shown as black dots in Fig.~\ref{fig:pivot}). 
Each of these contact points can have contacts with the ground and wall. 
Further, we considered a contact between the tip of the gripper and one side-face of the box.
The total number of the contact candidates was therefore 9.
Each contact candidate can take three states: (a) $\phi_i > 0$ (no contact), (b) $\phi_i = 0$ and $\psi_i \leq 0$ (sticking or sliding to the negative direction), and (c) $\phi_i = 0$ and $\psi_i > 0$ (sliding to the positive direction).
Therefore, the proposed method selects a contact mode from $3^9$ possible modes.

We designed the task space $\bm{g} (\bm{q})$ in the object function (\ref{eq:costOriginal}) as the angles of the gripper and box, i.e., $\bm{g} (\bm{q}) := [\theta_g \;\; \theta_b]^{\rm T}$ (see Fig.~\ref{fig:LCP_example}).
The initial and goal values of $(\theta_g, \theta_b)$ were set to $(0, \frac{\pi}{2})$ and $(\frac{\pi}{6}, 0)$, respectively. 

We conducted simulations under i.i.d. Gaussian noises with zero mean on the object shapes (i.e., modeling errors) and measurement states.
The modeling-error noises  were added to the length of each edge of the box with their standard deviation being: (the edge length) $\times$ (a model-error scale). 
Similarly, the measurement noises were added to the measured positions of the box and the gripper with the standard deviation being proportional to the object size.

We also evaluated the effectiveness of the relaxation of the complementarity constraints, by using Eqs.~(\ref{eq:normalLCP})--(\ref{eq:tangentialVelocityLCP})  instead of Eqs.~(\ref{eq:relaxedNormalLCP})--(\ref{eq:relaxedTangentialVelocityLCP}) for the LCQP in the simulations.

\subsubsection{Simulation results}
\begin{figure*}[tb]
\centering
\includegraphics[scale=0.45]{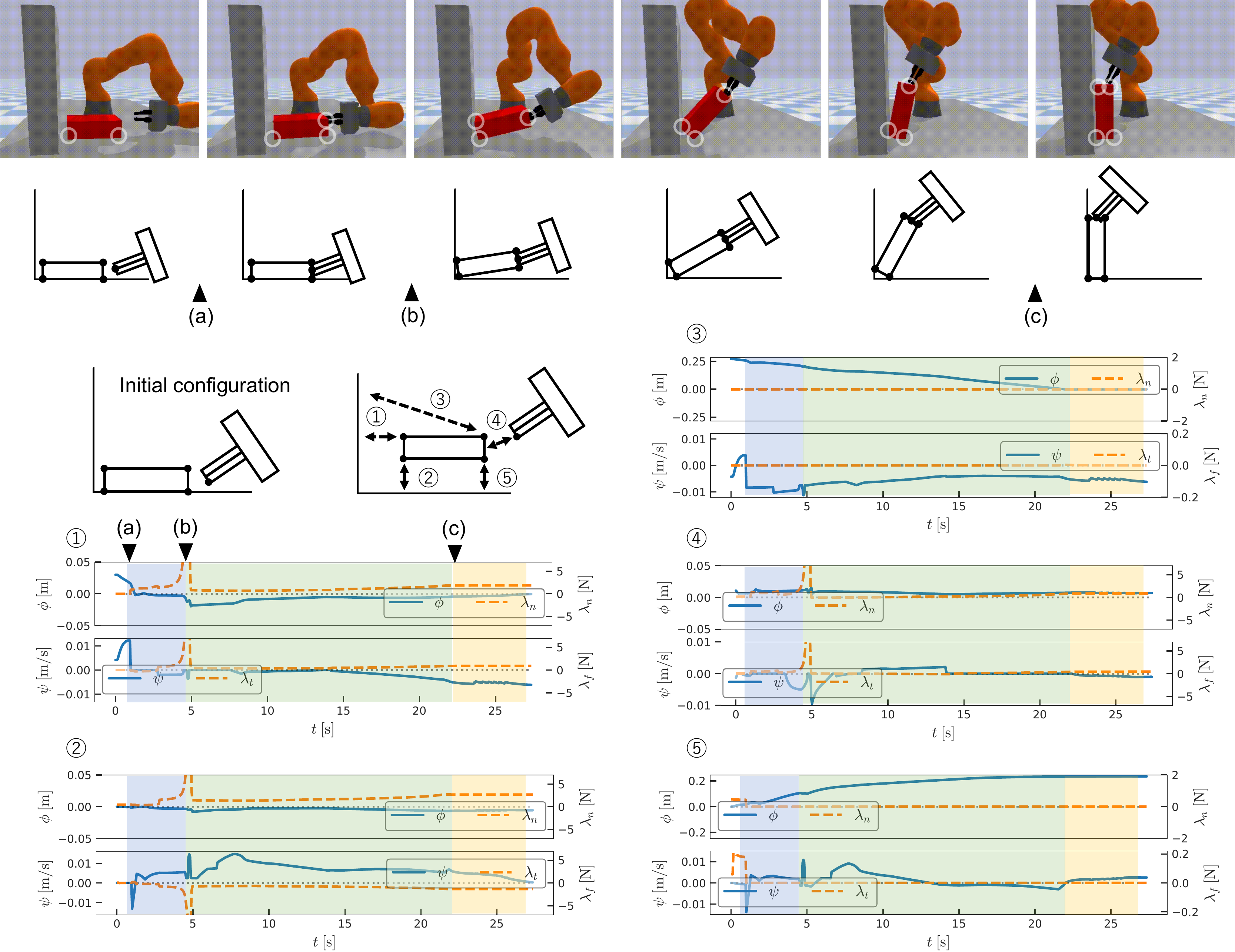}
\caption{3D dynamical simulation results of non-grasp pivoting of a box leveraging contacts with the environment (the ground and a wall) via the proposed LCQP controller. 
The top row shows the snapshots of the simulation in which white circles indicate the active contacts in the planar configuration.
The second top row shows the planar LCQP configurations corresponding to the snapshots.
Annotations (a), (b), and (c) indicate the switches of the contact states, which are only implicitly considered in the proposed LCQP controller. 
The five plots show the time series of the contact complementarities,  for five contact candidates, computed as the solutions of the proposed LCQP.}
\label{fig:snapshot}
\end{figure*}

Fig.~\ref{fig:snapshot} shows the results of the 3D dynamical simulations using the proposed LCQP controller without modeling and measurement errors. 
The top row shows snapshots of the  simulated manipulation process and the second row shows planar LCQP configurations corresponding to the snapshots.
These results show that the proposed method succeeded in pivoting the box by leveraging various contacts.
The contact state changed three times over the manipulation process, as annotated as (a), (b), and (c) in Fig. \ref{fig:snapshot}. Specifically, the box was lying on the ground at the beginning, with its two corners contacting with the ground (i.e., the state before (a) in Fig. \ref{fig:snapshot}).
Next, the gripper approached the box and pushed it, making a new contact between the gripper and the box (i.e., after the timing (a)). The gripper further pushed the box until it hit  the wall, making another new contact while breaking an existing contact from the ground (i.e., after the timing (b)).
The robot then kept the contact state and the force balance, until the box was finally stood up.

The five plots in Fig.~\ref{fig:snapshot} show the time series of contact complementarities computed for five contact candidates (see the annotations in Fig.~\ref{fig:snapshot}) as the solutions of the LCQP. In each plot, we show two contact complementarities in the top and bottom. In the top are the contact distance $\phi_i (\bm{q})$ and normal contact force $\lambda_{n_i}$, either of which should be zero. In the bottom are the tangential contact velocity $\psi_i (\bm{q}, \bm{v})$ and tangential contact force $\lambda_{f_i}$, whose signs should be opposite. 
These plots show that the proposed method computed the solution that almost satisfied these contact complementarities.
Meanwhile, the contact complementarities were not satisfied, e.g., when there are penetrations of objects due to numerical errors.
Nevertheless, the proposed LCQP controller successfully computed the feasible solutions thanks to relaxing the complementarity constraints.

Table~\ref{table:successRate} shows the success rates under different scales of modeling and measurement errors over 100 trials.
It shows that the proposed method can handle manipulation problems even under the model miss-matches and state estimation errors.
The LCQP controller without the relaxation never succeeded because of the numerical ill-conditioning in solving the LCQP due to  object penetrations, etc., showing the necessity for the relaxation of the complementarity constraints.

The LCQP solver took 7 ms on average and 20 ms at maximum. Therefore, the proposed controller achieved real-time computation of the continuous control action (gripper's position and force) while implicitly selecting contact modes.

\begin{table}[tb]
\centering
\caption{Success rates of non-grasip pivoting of a box with model errors and measurement noises over 100 trials}
\begin{tabular}{l|lllll}
\label{table:successRate}
\diagbox{Noise}{Model err.} & 0 &  $10 ^{-7}$ & $10 ^{-5}$ & $10 ^{-3}$ \\ \hline 
         0 & 1.00 & 0.9  & 0.64 & 0.57 \\
$10 ^{-7}$ & 0.99 & 0.78 & 0.75 & 0.59 \\
$10 ^{-5}$ & 0.7  & 0.78 & 0.78 & 0.58 \\
$10 ^{-3}$ & 0.6  & 0.58 & 0.57 & 0.48 \\
     0 (without relaxation) & 0 & 0 & 0 & 0 
\end{tabular}
\end{table}

\subsection{Various contact-rich manipulation tasks}

\begin{figure}[tb]
\centering
\includegraphics[scale=0.33]{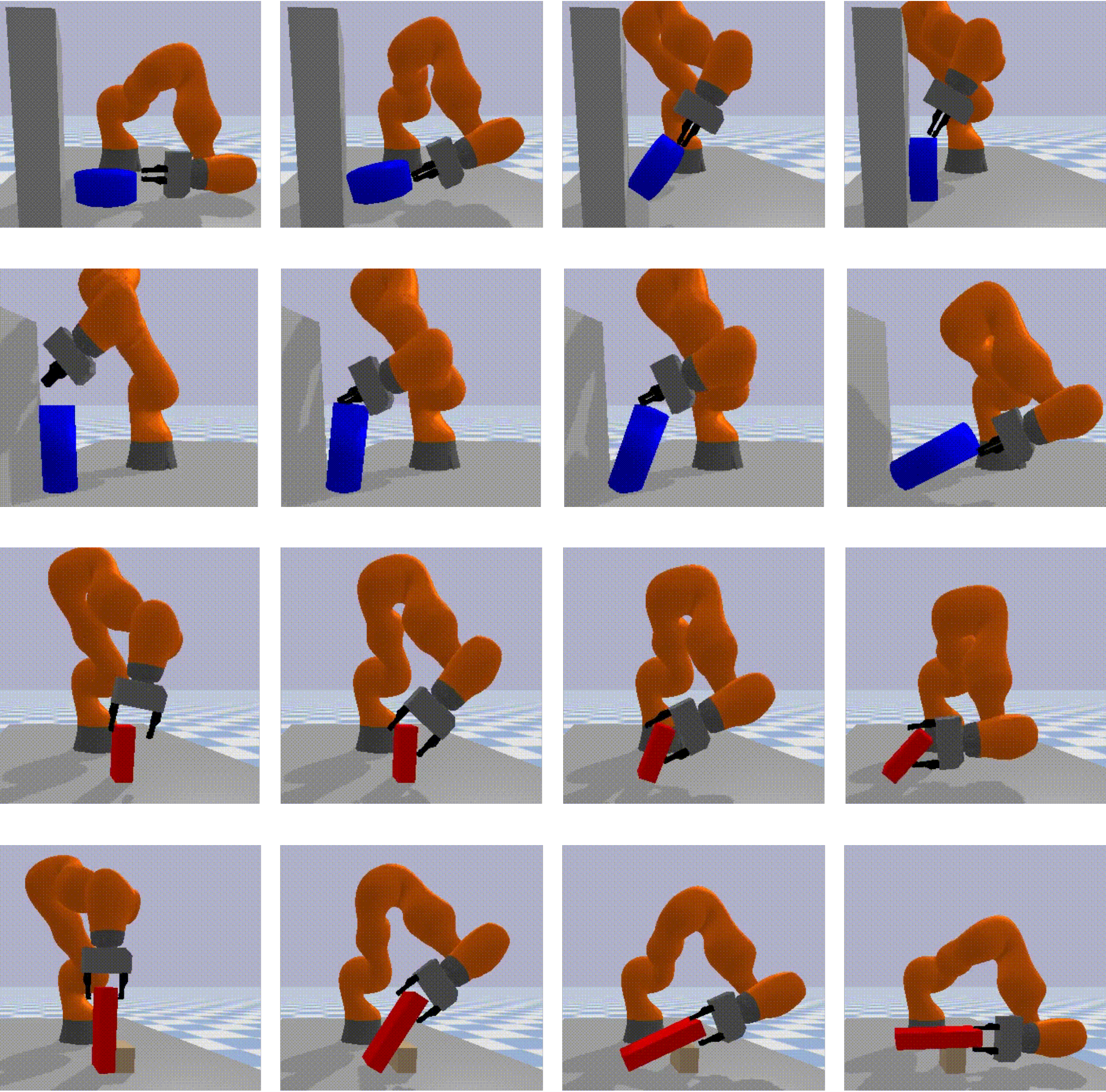}

\caption{Snapshots of 3D dynamical simulations of various contact-rich manipulation via the proposed LCQP controller.
First: pivoting a cylinder by leveraging the contacts with the ground and wall. 
Second: Pivoting to lay a cylinder. 
Third: Pivoting a box by using two fingers of the parallel gripper without strict grasping. 
Fourth: Pivoting a box by leveraging contacts with another box. 
}
\label{fig:varisousTasks}
\end{figure}

We further conducted 3D dynamical simulations for various contact-rich manipulation tasks, of which results are shown in Fig. \ref{fig:varisousTasks}.
The first task was non-grasp pivoting of a cylinder, whose planar LCQP formulation is identical to that of Sec.~\ref{subsection:example1}. The result thus shows that the LCQP formulation used in Sec~\ref{subsection:example1} also succeeded in this task.
The second task was to lay down a cylinder without grasping. 
The method also succeeded in this reverse action of the previous examples.
The third and fourth tasks were to pivot or lay down a box with loose grasping. Here, the fourth case further considers an additional box, which increases contact candidates.
These examples are different from the previous examples in that they use both fingers of the parallel gripper.
We thus considered additional contact candidates regarding the two finger tips and the surfaces of fingers, as illustrated in Fig.~\ref{fig:LCP_example}.
Our method succeeded in both tasks. These results show the versatility of the proposed framework.

\section{Conclusions}\label{section:Conclusion}
We have proposed a versatile local planning and control framework for contact-rich manipulation that determines the continuous control action under the various contact modes online.
We modeled the physical characteristic of contact-rich manipulation via quasistatic dynamics and complementarity constraints.
We formulated an LCQP with these constraints so that online execution is affordable.
We conducted dynamical simulations on a 3D physical simulator and demonstrated that the proposed method can achieve various contact-rich tasks by online computation of the control actions including contact modes. 
Meanwhile, improving the robustness of the proposed method is a part of important future work.  
This includes an extension of the current simple joint controller in Fig. \ref{fig:control_loop} to more sophisticated methods.
Future works also include an extension of the 2D LCQP formulation to 3D and combination with object pose and parameters (shapes, mass, etc.) estimation.







\bibliographystyle{IEEEtran}
\bibliography{IEEEabrv, ieee}


\end{document}